\documentclass[aps,pre,twocolumn,superscriptaddress,nofootinbib,floatfix]{revtex4-2}
\usepackage{amsmath,amssymb}
\usepackage{bm}
\usepackage{graphicx}
\usepackage[colorlinks=true,allcolors=blue]{hyperref}

\newcommand{\dd}{\mathrm{d}}
\newcommand{\Lag}{\mathcal{L}}
\newcommand{\Action}{\mathcal{A}}
\newcommand{\pth}{\Gamma}
\newcommand{\rpath}{\widetilde{\Gamma}}
\newcommand{\z}{\bm{z}}
\newcommand{\f}{\bm{f}}
\newcommand{\vv}{\bm{v}}
\newcommand{\Dif}{D}
\newcommand{\Hess}{\mathcal{H}}

\begin{document}

\title{A Path-Space Formulation of Prediction in World Models:
From a Single Action to Prediction, Planning, and Irreversibility}

\author{Gunn Kim}
\affiliation{Department of Physics, Sejong University, Seoul 05006, Republic of Korea}
\date{\today}

\begin{abstract}
We propose a path-space formulation of prediction in AI world models. Rather than sequences of one-step conditional distributions, we argue that a world model implicitly defines a probability measure over future trajectories. In the local regime where latent dynamics admit an effective Markovian description, this path measure takes the Onsager–Machlup form. Within this framework, prediction (most probable trajectory), planning (constrained optimization), and uncertainty (fluctuations) emerge as operations on a single action functional. We decompose the latent dynamics into reversible and irreversible components and introduce operational measures of entropy production from model rollouts. In controlled small-scale attention-based models, we find that attention asymmetry is acquired during training in proportion to the irreversibility of the data. Symmetrizing the learned attention suppresses entropy production and selectively degrades long-horizon prediction of irreversible dynamics while preserving relaxational prediction. These results suggest that irreversibility may serve as a computational resource for predictive world models. More generally, the fundamental predictive object is a distribution over future paths rather than states.
\end{abstract}

\maketitle

\section{Introduction}
\label{sec:intro}

A world model is a learned internal simulator of an environment: given a context,
it generates futures consistent with the data on which it was trained~\cite{ha2018,hafner2023}. The idea recurs across disciplines. In neuroscience, the cortex is described as a prediction machine that continually anticipates its sensory stream~\cite{friston2010}; in machine learning, sequence models and latent world models are trained to roll an environment
forward~\cite{vaswani2017,hafner2023}. In all of these settings, a predictive system is summarized by the same slogan, that it is a ``future-prediction machine,'' and that slogan is almost always made precise through the one-step conditional
$p(x_{t+1}\mid x_t)$, the quantity that autoregressive training directly optimizes.

We take a different basic object. Prediction, planning, and uncertainty are all
assertions about futures, that is, about whole trajectories rather than single
increments, and the natural carrier of all three is the probability measure that
the model assigns to future paths. Writing a future as a latent trajectory
$\pth=\{\z(t)\}$, we study the functional measure $P[\pth]$ that the model induces
on such trajectories; the one-step conditional, and the next token it produces are then derived marginals of $P[\pth]$. The questions of what a world model predicts, how it plans, and how confident it is then become questions about the structure of a
single path distribution. That the fundamental predictive object of a world model
is a distribution over future paths rather than future states is the conceptual
core of this work.

This shift is useful because it imports a mature physical machinery. In the regime
where the learned latent dynamics is effectively local in time, the Markovian limit
of its Mori--Zwanzig formalism~\cite{mori1965,zwanzig2001,chorin2000}, the
path measure takes the Onsager--Machlup form $P[\pth]\propto
e^{-\Action[\pth]}$~\cite{onsager1953}, the action functional of a diffusion. A
single functional then governs three operations that are usually treated as
separate modules: prediction is its most-probable path, planning is its constrained
least-action path, and predictive uncertainty is its curvature. The same action
separates the drift into a reversible, gradient part and an irreversible,
circulating part, so that the entropy production of nonequilibrium statistical
mechanics~\cite{seifert2012} becomes a measurable property of the predicted world
rather than an abstract thermodynamic quantity. The framework thereby provides a
common language linking latent dynamics, stochastic thermodynamics, and
attention-based sequence modeling.

Within this language the architecture acquires a thermodynamic reading. We show how an attention layer can be read as computing the local action, with the query--key product $W_Q^\top W_K$ playing the role of the metric of the kinetic term and its
antisymmetric part that of the irreversible drift. An explicit calculation of the drift Jacobian then identifies the query-key asymmetry as an architecturally controllable source of irreversibility. These statements are operational: from
model rollouts we estimate the drift, the circulating current, the entropy
production, the non-normality, and the attention asymmetry, and we state a list of
falsifiable predictions. In a small trained attention model these quantities are measurements rather than assumptions. The attention asymmetry, the entropy production, and the non-normality are acquired together during learning in proportion to the irreversibility of the data; a causal intervention that symmetrizes the attention collapses all three. The same intervention selectively destroys long-horizon prediction of circulating structure while sparing relaxational prediction. Irreversibility is in this sense not only a by-product of learning but a measurable resource for prediction.

To delimit our contribution precisely: our primary claim does not concern the architectural necessity of attention in world models. Rather, we establish that predictive dynamics are fundamentally structured as inference on path space, where temporal irreversibility manifests as a measurable computational resource. The drift-Jacobian calculation identifies and lets us
control one channel of irreversibility. Whether that channel dominates the entropy
production of a large pretrained architecture, in which residual connections,
feedforward blocks, normalization, value mixing, and depth all contribute to the
learned drift, is a separate empirical question that we leave open. In the
tradition where a minimal model is used to exhibit a universal mechanism rather
than to simulate a system at full scale, we develop and test the structure in a
controlled, small-scale setting.

The paper is organized as follows. Section~\ref{sec:pathspace} formalizes the path
measure and its local Onsager--Machlup action. Section~\ref{sec:prediction} treats
prediction as the most-probable future and identifies where it departs from the
deterministic rollout. Section~\ref{sec:planning} develops planning and uncertainty
as further operations on the same action. Section~\ref{sec:attention} establishes
the correspondence between attention and the local action and derives the
antisymmetric Jacobian from the query--key asymmetry. Section~\ref{sec:irreversibility}
gives operational definitions of the reversible and irreversible drift, the entropy
production, and the attention asymmetry, and states falsifiable predictions.
Section~\ref{sec:results} presents the controlled experiments and discusses their
scope, and the final section concludes.

\section{The predictive object: world models on path space}
\label{sec:pathspace}

\subsection{From next-token conditionals to future trajectories}

Let an encoder map observations to a latent state $\z\in\mathbb{R}^d$, in which
the model's dynamics are defined. A future is a latent trajectory
\begin{equation}
  \pth=\{\z(t)\}_{0\le t\le T},
\end{equation}
and the predictive object we study is the functional probability measure
$P[\pth]$ that the model induces on such trajectories given a context. The
one-step conditional, and with it the next token, is recovered as a marginal of
$P[\pth]$ and is in this sense a derived rather than a fundamental quantity. The
central questions of this paper, what the model predicts, how it plans, and how
confident it is, become questions about the structure of $P[\pth]$. Of the three
operations on $P[\pth]$ that follow, it is prediction, and specifically its
irreversible part, that we make operational and test; planning and uncertainty are
structural consequences of the same action, developed here for completeness and
left to direct measurement elsewhere.

\subsection{The latent dynamics and its local regime}
\label{sec:local}

We model the latent evolution as a stochastic process
\begin{equation}
  \dd\z = \f(\z)\,\dd t + \sqrt{2\Dif}\,\dd \bm{W}_t,
  \label{eq:sde}
\end{equation}
with drift $\f$ and, for clarity, isotropic and constant diffusion $\Dif$; the
anisotropic case $\Dif(\z)$ introduces only the standard multiplicative-noise
corrections and is deferred. Two points of principle attach to
Eq.~\eqref{eq:sde}, since a trained world model is in general neither Markovian
nor continuous in time.

First, the exact reduced dynamics obtained after integrating out the latent
coordinates that the model does not expose is, by the Mori--Zwanzig
projection~\cite{mori1965,zwanzig2001,chorin2000},
a \emph{generalized} Langevin equation with a memory kernel and colored noise,
\begin{equation}
\begin{aligned}
  \dot\z(t) &= \f_{\mathrm{loc}}(\z) - \int_0^t K(t-t')\,\dot\z(t')\,\dd t'
              + \bm{\xi}(t), \\
  &\quad \langle\bm{\xi}(t)\bm{\xi}(t')\rangle \propto K(t-t').
\end{aligned}
  \label{eq:gle}
\end{equation}
The local model \eqref{eq:sde} is the leading term of \eqref{eq:gle} in the
Markovian limit $K(t)\to 2\gamma\,\delta(t)$, valid when the memory time
$\tau_{\mathrm{mem}}$ (the width of $K$, set physically by the attention's
effective look-back) is short compared with the dynamical time $\tau_{\mathrm{dyn}}$
on which we predict. We make this restriction explicit and treat the small
parameter
\begin{equation}
  \epsilon \equiv \tau_{\mathrm{mem}}/\tau_{\mathrm{dyn}}
\end{equation}
as the control parameter of the expansion: the local theory is its
$\epsilon\to 0$ limit, and non-local, higher-time-derivative corrections are
organized in powers of $\epsilon$. This is the working analogue of Hooke's law.
A real spring has an elastic limit, but the linear regime is a controlled and
honest description once its range of validity is stated and, ideally, measured.
We work throughout in this regime and report $\epsilon$ rather than assuming it
away. The memory time itself is architectural: it is bounded above by the context
window, $\tau_{\mathrm{mem}}\le N\Delta t$ for context length $N$, and grows with
depth $L$, since each layer composes attention over the already-mixed history of
the previous one, so that $\epsilon$ is set by $N$ and $L$ relative to
$\tau_{\mathrm{dyn}}$. The local regime is the one in which the learned attention
is effectively short-ranged on the dynamical timescale, a condition that is itself
measurable (P1 below).

Second, going local does not mean going to equilibrium. We retain the full
drift, including its non-gradient part. Writing
\begin{equation}
  \f = -\nabla U + \vv,
  \label{eq:helmholtz}
\end{equation}
the gradient part $-\nabla U$ is the reversible, detailed-balance-respecting
component, while the circulating part $\vv$, defined by
$\nabla\!\cdot(\rho_{\mathrm{ss}}\vv)=0$ with $\rho_{\mathrm{ss}}$ the stationary
density, breaks detailed balance and produces entropy. The local, linear
elastic regime that still carries an antisymmetric, non-reciprocal response is
exactly the setting of odd elasticity, in which the symmetric part of the
stiffness plays the role of $-\nabla U$ and the antisymmetric (odd) part plays
the role of $\vv$. Restricting to the local regime therefore preserves, rather
than discards, the irreversible structure that the rest of the paper depends on.

\subsection{The path measure and its Onsager--Machlup action}
\label{sec:om}

For the local dynamics \eqref{eq:sde}, the probability that the trajectory lies
in an infinitesimal tube around a given path $\pth$ is, up to a
path-independent normalization, $P[\pth]\propto e^{-\Action[\pth]}$, with the
Onsager--Machlup action~\cite{onsager1953}
\begin{equation}
  \Action[\pth] = \int_0^T \dd t\,\Lag(\z,\dot\z),
  \qquad
  \Lag = \frac{1}{4\Dif}\,\bigl|\dot\z-\f(\z)\bigr|^2
         + \frac{1}{2}\,\nabla\!\cdot\f .
  \label{eq:OMaction}
\end{equation}
The first term is the kinetic weight that penalizes deviation of the realized
velocity from the drift; the second is the Jacobian term arising from the
Stratonovich (midpoint) discretization of \eqref{eq:sde}, and it is the term
that distinguishes the genuine path probability from the naive squared-residual
cost. The normalization is the path integral
\begin{equation}
  P[\pth]=\frac{1}{Z}\,e^{-\Action[\pth]},
  \qquad
  Z=\int \mathcal{D}\pth\; e^{-\Action[\pth]} .
  \label{eq:gibbs}
\end{equation}
Equation~\eqref{eq:gibbs} is formally a Gibbs measure on path space, with
$\Action$ playing the role of an energy and the diffusion $\Dif$ setting the
temperature scale. Two structures we shall use repeatedly follow at once. The
action is additive along time, $\Action=\int\dd t\,\Lag$, so that the
log-probability of a future factorizes into local-in-time increments, which is
the precise content of the locality assumption of Sec.~\ref{sec:local}. And
$\Action$ splits into a time-symmetric and a time-antisymmetric part under the
path reversal $\pth=\{\z(t)\}\mapsto\rpath=\{\z(T-t)\}$,
\begin{equation}
  \Action = \Action_{\mathrm{sym}} + \Action_{\mathrm{irr}},
  \quad
  \Action_{\mathrm{irr}}[\pth]
  = -\frac{1}{2\Dif}\int_0^T \vv(\z)\cdot \dd\z ,
  \label{eq:split}
\end{equation}
so that the irreversible part of the action is the line integral of the
circulating drift along the path. The path-wise entropy production is the
log-ratio of forward and reversed path probabilities,
\begin{equation}
  \Sigma[\pth] = \ln\frac{P[\pth]}{P[\rpath]}
  = \frac{1}{\Dif}\int_0^T \vv(\z)\cdot\dd\z ,
  \label{eq:entropy}
\end{equation}
a quantity that is defined directly from the path measure and its reversal and,
as we stress later, does not itself require the local form \eqref{eq:OMaction}.

\subsection{One functional, three operations}

The point of Eq.~\eqref{eq:gibbs} is that a single functional organizes the three
operations a world model must perform. They are developed in
Secs.~\ref{sec:prediction}--\ref{sec:attention}; we state them here to fix the
logic of the paper.
\begin{itemize}
  \item \emph{Prediction} is the stationary, most-probable future,
        $\pth^\ast=\arg\min_\pth \Action[\pth]$, obtained from
        $\delta\Action=0$. We show in Sec.~\ref{sec:prediction} that
        $\pth^\ast$ is generically distinct from the deterministic rollout
        $\dot\z=\f$, and that the circulating drift $\vv$ enters it as a
        Lorentz-like force.
  \item \emph{Planning} is the least-action future subject to terminal
        constraints, with the planning value given by the path free energy
        $-\ln\!\int_{\,\z_0\to\,\mathrm{goal}}\mathcal{D}\pth\,e^{-\Action}$.
        Prediction and planning are thus the free- and fixed-endpoint versions
        of one variational problem.
  \item \emph{Uncertainty} is the curvature of the action about $\pth^\ast$.
        Expanding $\Action=\Action[\pth^\ast]+\tfrac12\langle\eta,\Hess\eta\rangle+\cdots$
        with $\Hess=\delta^2\Action/\delta\pth^2$, the predictive covariance is
        $\Hess^{-1}$, the Green's function of a Schr\"odinger-type fluctuation
        operator along the path.
\end{itemize}
The reversible/irreversible split \eqref{eq:split} threads all three, and
Sec.~\ref{sec:irreversibility} connects its irreversible part to the query--key
asymmetry of attention.

\subsection{Relation to prior work}

Each ingredient of the framework has a counterpart in the literature, and we state
the correspondences plainly so that the contribution is not mistaken for any one of
them. The static, single-step reading of \eqref{eq:gibbs}, a Gibbs distribution
over candidate continuations with energy equal to the negative attention score, is
the modern Hopfield-network view of attention~\cite{ramsauer2020}. The
fixed-endpoint, planning reading is control-as-inference, in which optimal
behavior is a trajectory distribution weighted by return~\cite{levine2018}. The
use of an Onsager--Machlup action for a learned generative process is established
for diffusion and flow-matching
models~\cite{song2021,lipman2023}, where the learned score is the
drift~\cite{raja2025}. The locality regime rests on the Mori--Zwanzig projection
that produces a generalized Langevin equation~\cite{zwanzig2001}, and the
odd-elastic reading of the antisymmetric drift is borrowed from non-reciprocal
continuum mechanics~\cite{scheibner2020}.

Two further lines are close enough to require explicit separation. The
dynamical-systems analysis of trained recurrent networks reverse-engineers their
computation from fixed points and the linearized flow around
them~\cite{sussillo2013}, and a parallel line reads the transformer itself as an
interacting particle system whose tokens flow and cluster under
attention~\cite{geshkovski2023}; these are the nearest precedents to our reading
of a world model through its drift field, but they characterize geometry, fixed
points and slow manifolds, not thermodynamics, and they have no attention
parameter to trace the geometry back to. Stochastic thermodynamics of neural and
brain dynamics measures
entropy production and its oscillatory decomposition from recorded
activity~\cite{lynn2021,sekizawa2024}; this supplies the irreversibility
diagnostic we use, but on data, with no handle on the architecture that produced
it.

What is new is therefore not any single correspondence but their unification and
its consequence. The unification reads prediction, planning, and uncertainty as
operations on one action that attention computes. The consequence is the bridge,
absent from all of the above, from an architectural quantity, the query--key
asymmetry, to a thermodynamic one, the entropy production of the predicted
dynamics, and from there to a functional one: that this entropy production is a
resource the model spends to predict irreversible structure over a long horizon
(Sec.~\ref{sec:fig3}). The geometric and thermodynamic literatures measure such
quantities in recorded dynamics; here they are produced, controlled, and shown to
matter for prediction by a single architectural knob.

\section{Prediction: the most-probable future}
\label{sec:prediction}

Prediction is the operation $\delta\Action=0$ on the path action of
Sec.~\ref{sec:om}. Carrying it out gives two facts that are easy to state and
easy to forget: the most-probable future obeys a Newtonian equation in which the
irreversible part of the drift acts as a magnetic force, and that future is
generically not the deterministic rollout $\dot\z=\f$.

\subsection{The Euler--Lagrange equation}

Stationarity of $\Action[\pth]=\int_0^T\Lag\,\dd t$ with the Onsager--Machlup
Lagrangian \eqref{eq:OMaction}, through
$\tfrac{\dd}{\dd t}\,\partial_{\dot\z}\Lag=\partial_{\z}\Lag$, and using the
identities $\partial_j f_i-\partial_i f_j=2(J_A)_{ij}$ and
$f_j\partial_i f_j=\tfrac12\partial_i|\f|^2$, gives for constant isotropic $\Dif$
\begin{equation}
  \ddot\z = 2\,J_A(\z)\,\dot\z \;-\; \nabla V_{\mathrm{eff}}(\z),
  \quad
  V_{\mathrm{eff}} = -\tfrac12|\f|^2 - \Dif\,\nabla\!\cdot\f,
  \label{eq:instanton}
\end{equation}
with $J_{ij}=\partial_j f_i$ the drift Jacobian and $J_A=\tfrac12(J-J^\top)$ its
antisymmetric part. Equation~\eqref{eq:instanton} is the instanton equation of
the path measure: the most-probable future is a Newtonian trajectory in the
effective potential $V_{\mathrm{eff}}$ under an additional velocity-dependent
force $2J_A\dot\z$. Because $J_A$ is antisymmetric this force does no work,
$\dot\z\cdot(2J_A\dot\z)=0$; it is a magnetic, or gyroscopic, force, with $2J_A$
in the role of the field-strength tensor. The Lagrangian has no explicit time, so
\begin{equation}
  E = \frac{1}{4\Dif}\,|\dot\z|^2 + \frac{1}{2\Dif}\,V_{\mathrm{eff}}(\z)
\end{equation}
is conserved along the most-probable path, a first integral of
Eq.~\eqref{eq:instanton}.

\subsection{Why the most-probable future is not the deterministic rollout}

It is tempting to identify the predicted future with the deterministic flow
$\dot\z=\f$, the path that makes the kinetic term of $\Action$ vanish. This is
correct only in a special case. Substituting $\dot\z=\f$ into
Eq.~\eqref{eq:instanton} and using $2J_A\f+\nabla(\tfrac12|\f|^2)=J\f$, the
deterministic flow solves the Euler--Lagrange equation if and only if
\begin{equation}
  \Dif\,\nabla(\nabla\!\cdot\f) = \bm{0}.
  \label{eq:rolloutcond}
\end{equation}
Whenever the drift has non-uniform divergence, the most-probable future departs
from the deterministic rollout, and the leading departure is the noise-induced
term $\Dif\,\nabla(\nabla\!\cdot\f)$, of order $\Dif$. The interpretation is
standard but worth stating: integrating the learned drift forward returns the
mean field, not the mode of the future distribution, and the two differ wherever
probability is being focused or defocused ($\nabla\!\cdot\f\neq\mathrm{const}$). A
world model that reports its single most-likely rollout is therefore not
reporting $\arg\max_\pth P[\pth]$ unless its drift is divergence-harmonic.

\subsection{The antisymmetric Jacobian threads prediction and irreversibility}

The gyroscopic force in Eq.~\eqref{eq:instanton} is generated by $J_A$, the same
antisymmetric Jacobian that in Sec.~\ref{sec:irreversibility} sources the
circulating drift $\vv$ and the entropy production, and that in
Sec.~\ref{sec:attention} descends from the query--key asymmetry of attention. One
object thus controls both the geometry of prediction and its thermodynamics. On
the forward, free-endpoint path its effect is mild, entering only through the
off-deterministic corrections above. It becomes decisive in the two problems
treated next: the fixed-endpoint problem of planning
(Sec.~\ref{sec:planning}), where the instanton must reach a prescribed target and
the magnetic force bends it away from any gradient descent of $V_{\mathrm{eff}}$,
and the fluctuation spectrum that sets predictive uncertainty, where $J_A$ enters
the Hessian. A world model with symmetric attention ($J_A=0$) has the curl-free
instanton $\ddot\z=-\nabla V_{\mathrm{eff}}$ and predicts only gradient
relaxation; the circulating, history-carrying futures require $J_A\neq0$.

\section{Planning and uncertainty from the same action}
\label{sec:planning}

Prediction fixed the initial condition and left the future free. Planning fixes a
target and asks for the best route to it; uncertainty asks how sharply that route
is determined. Both are read off the same path integral, by imposing a terminal
condition and by expanding to second order.

\subsection{Planning is fixed-endpoint least action}

Given a target, a terminal state $\z_T$ or more generally a terminal cost, the
optimal plan is the least-action path that reaches it,
\begin{equation}
  \pth^\ast=\arg\min_{\z(0)=\z_0,\;\z(T)=\z_T}\Action[\pth],
\end{equation}
the instanton \eqref{eq:instanton} now solved as a boundary-value problem. The
gyroscopic force $2J_A\dot\z$, mild for free-endpoint prediction, here bends the
plan away from gradient descent of $V_{\mathrm{eff}}$: an asymmetric attention
plans along curved, circulating routes, a symmetric one only down the potential.
The value of the target is the path free energy,
\begin{align}
  \mathcal{V}(\z_T,T\mid\z_0)
  &= -\Dif\ln\!\!\int_{\z_0\to\z_T}\!\!\mathcal{D}\pth\;e^{-\Action[\pth]}\nonumber\\
  &= -\Dif\ln K(\z_T,T\mid\z_0,0),
\end{align}
with $K$ the propagator. In the low-noise limit
$\mathcal{V}\to\Dif\,\Action[\pth^\ast]$, and $\mathcal{V}$ obeys a
Hamilton--Jacobi--Bellman equation whose characteriztics are the instantons. This
is the control-as-inference correspondence~\cite{levine2018}, here grounded in the
Onsager--Machlup action: the optimal cost-to-go is the free energy of the future
ensemble, and planning is least action under a terminal constraint.

\subsection{Value and uncertainty are consecutive orders of one expansion}

Expanding the path integral about $\pth^\ast$ by the saddle point gives
\begin{equation}
  -\ln\!\int\!\mathcal{D}\pth\,e^{-\Action}
  = \Action[\pth^\ast] + \tfrac12\ln\det\Hess + O(\Dif),
  \quad
  \Hess = \frac{\delta^2\Action}{\delta\pth^2}\bigg|_{\pth^\ast}.
\end{equation}
The leading term is the planning value; the next, the fluctuation determinant, is
the log-volume of nearby futures, that is the predictive uncertainty. Prediction,
planning, and uncertainty are thus not three constructions but three terms read
off one functional: the stationary path, its action, and its curvature.

\subsection{The fluctuation operator and the predictive covariance}

The second variation is the quadratic form
$\tfrac12\int_0^T\eta^\top\Hess\,\eta\,\dd t$ with the Jacobi operator
\begin{equation}
  \Hess = \frac{1}{2\Dif}\Big[-\frac{\dd^2}{\dd t^2}
          + 2J_A(t)\,\frac{\dd}{\dd t} + \mathsf{\Omega}^2(t)\Big],
  \label{eq:hessop}
\end{equation}
where $\mathsf{\Omega}^2(t)$ is a symmetric matrix potential assembled from
$J^\top J$, $\dot J_S$, and the curvature $\nabla\nabla V_{\mathrm{eff}}$ along
$\pth^\ast$, and $J_S=\tfrac12(J+J^\top)$. The boundary conditions complete the
definition: Dirichlet $\eta(0)=\eta(T)=\bm0$ for the fixed-endpoint planning
covariance, and $\eta(0)=\bm0$ with the natural condition
$\partial_{\dot\eta}\Lag=\bm0$ at $t=T$ for forward prediction. With these
conditions $\Hess$ is self-adjoint and, absent a conjugate point, invertible, so
the predictive covariance is its Green's function,
\begin{equation}
\begin{split}
  \langle\eta(t)\,\eta(t')^\top\rangle &= \Hess^{-1}(t,t')\equiv G(t,t'),\\
  \Hess\,G(t,t') &= \mathbb{1}\,\delta(t-t').
\end{split}
\end{equation}
The antisymmetric Jacobian enters the first-order term $2J_A\partial_t$, so a
world model with asymmetric attention has a non-normal fluctuation operator: its
predictive uncertainty grows transiently and anisotropically rather than
diffusing isotropically, the same non-normality that drives the entropy
production of Sec.~\ref{sec:irreversibility}. Symmetric attention ($J_A=0$)
returns a normal operator and isotropic, monotone uncertainty growth.

\section{Attention computes the local action}
\label{sec:attention}

This section is the architectural bridge. It reads the autoregressive
attention computation as a discretization of the path integral of
Sec.~\ref{sec:om}, so that the attention logits play the role of local action
increments and both terms of the Onsager--Machlup Lagrangian can be read off from
the attention operation. We present this as a structural correspondence rather than
a derivation from first principles. It holds at the level of a single attention
layer in the local regime, where the one-step conditional is Gaussian; a deep stack
renormalizes the resulting drift and diffusion, which we return to in
Sec.~\ref{sec:results}.

\subsection{Autoregressive attention is a discretized path measure}

An autoregressive world model generates a trajectory by sampling each step from a
conditional built by attention. The joint law of the generated path factorizes,
\begin{equation}
\begin{aligned}
  -\ln P[\pth] &= \sum_t \big[-\ln p(\z_{t+1}\mid\z_{\le t})\big] \\
  &\xrightarrow[\text{local}]{} \sum_t \ell(\z_t,\z_{t+1}) \\
  &\xrightarrow[\Delta t\to0]{} \int_0^T \Lag(\z,\dot\z)\,\dd t ,
\end{aligned}
\end{equation}
where the middle arrow is the locality (Markov) assumption of
Sec.~\ref{sec:local} and the last is the continuum limit. Because the one-step
conditional is Gaussian with mean $\z_t+\Delta t\,\f$ set by the attention readout,
its negative logarithm is exactly the discretized Onsager--Machlup increment
$\ell(\z_t,\z_{t+1})=\tfrac{1}{4\Dif\Delta t}\,|\z_{t+1}-\z_t-\Delta t\,\f|^2
+\tfrac12\nabla\!\cdot\f$; the attention enters the action through the drift $\f$,
not as the increment itself, and the two terms of the Lagrangian are read off from
$\f$ in the next subsections.

\subsection{Drift, temperature, and metric from the attention layer}

A single attention layer reads out
\begin{equation}
\begin{aligned}
  \mathrm{Attn}(\z) &= \sum_s a_s(\z)\,\bm{v}_s,\qquad
  a_s(\z) = \frac{e^{\,s_s(\z)}}{Z(\z)}, \\
  s_s(\z) &= \frac{1}{\sqrt d}\,\z^\top M\,\bm{m}_s,\qquad
  M=W_Q^\top W_K,
\end{aligned}
\end{equation}
over context keys $\bm{m}_s$ and values $\bm{v}_s$, with denominator
$Z(\z)=\sum_s e^{s_s(\z)}$. The one-step update
$\z_{t+1}=\z_t+\Delta t\,\f(\z_t)+\sqrt{2\Dif}\,\Delta\bm W$, with
$\f(\z)=\mathrm{Attn}(\z)-\z$, then fixes three objects of the action. The inverse
temperature weighting the path Gibbs measure is the attention scale $\sqrt d$,
the Lagrange multiplier of Sec.~\ref{sec:om}. The bilinear form $M=W_Q^\top W_K$
is the metric of the kinetic term, so a non-symmetric $M$ is an anisotropic,
non-reciprocal metric. Its antisymmetric part $M_A$ feeds the antisymmetric drift
Jacobian $J_A$, derived explicitly in Sec.~\ref{sec:jacobian-asym}, and through
Eqs.~\eqref{eq:instanton} and \eqref{eq:vmeasure} the same $M_A$ supplies both the
gyroscopic force of prediction and the entropy production of
Sec.~\ref{sec:irreversibility}. In the symmetric, value-tied limit ($W_Q=W_K$
with $\bm{v}_s$ tied to $\bm{m}_s$) the readout reduces to a gradient flow,
recovering the modern Hopfield energy and a purely relaxational world model.

\subsection{The Jacobian term from the softmax denominator}

The second term of the Onsager--Machlup Lagrangian, $\tfrac12\nabla\!\cdot\f$, is
the Jacobian of the midpoint (Stratonovich) discretization, and it is what
distinguishes the path probability from a bare squared residual. In a transformer
it is supplied by the softmax denominator. Differentiating the normalized weights
gives $\nabla a_s=\tfrac{1}{\sqrt d}\,a_s\,(M\bm{m}_s-\langle M\bm{m}\rangle_a)$
with $\langle\cdot\rangle_a=\sum_s a_s(\cdot)$, so that
\begin{equation}
\begin{split}
  \tfrac12\,\nabla\!\cdot\f
  &= \frac{1}{2\sqrt d}\sum_s a_s\,(M\bm{m}_s-\langle M\bm{m}\rangle_a)\cdot\bm{v}_s\\
  &= \frac{1}{2\sqrt d}\,\mathrm{Cov}_a\!\big(M\bm{m},\,\bm{v}\big).
\end{split}
\label{eq:jacfromsoftmax}
\end{equation}
The Jacobian term is thus the attention-weighted covariance, under the softmax
denominator, between the key projection $M\bm{m}_s$ and the value $\bm{v}_s$. Both
pieces of the local action therefore come from one attention layer: the kinetic
term from the score, the Jacobian term from the normalization. Key-normalization
(LayerNorm) enters by making the score a genuine inner product and fixing the
temperature scale, the same role it plays in reducing free diffusion to
dot-product attention. Equation~\eqref{eq:jacfromsoftmax} also makes the locality
check of Sec.~\ref{sec:irreversibility} concrete: it holds layer by layer, and the
departure from it across a deep stack measures the memory parameter $\epsilon$.

\subsection{The antisymmetric Jacobian descends from the query--key asymmetry}
\label{sec:jacobian-asym}

The paper's causal claim, that the query--key asymmetry is an architecturally
controllable source of the irreversible drift, can be made explicit at the level of
the drift Jacobian rather than asserted. Using $\nabla a_s$ from above, the Jacobian
of $\f=\mathrm{Attn}(\z)-\z$ is the attention-weighted cross-covariance between the
values and the key projections,
\begin{equation}
\begin{split}
  J &= \frac{1}{\sqrt d}\,\mathrm{Cov}_a\!\big(\bm{v},\,M\bm{m}\big) - I ,\\
  \mathrm{Cov}_a(\bm{v},\bm{u}) &= \langle\bm{v}\bm{u}^\top\rangle_a
   -\langle\bm{v}\rangle_a\langle\bm{u}\rangle_a^\top .
\end{split}
\end{equation}
In the value-tied regime $\bm{v}_s=\bm{m}_s$ of Sec.~\ref{sec:attention}, where the
query--key product is the only asymmetry channel, this cross-covariance is
$C\,M^\top$ with $C=\mathrm{Cov}_a(\bm{m},\bm{m})$ the symmetric, positive
attention-weighted key covariance, and the antisymmetric part of the Jacobian
splits into exactly two terms,
\begin{equation}
  J_A = \frac{1}{2\sqrt d}\Big(
        \underbrace{[\,C,\,M_S\,]}_{\text{state-dependent}}
        \;-\;
        \underbrace{\{\,C,\,M_A\,\}}_{\text{architectural}}\Big).
  \label{eq:JAfromMA}
\end{equation}
The second term is the anticommutator of the key covariance with the query--key asymmetry $M_A$. For a positive-definite $C$, it vanishes if and only if $M_A=0$, so $M_A$ is exactly the architecturally controlled source of the antisymmetric Jacobian, and hence, through Eq.~\eqref{eq:vmeasure}, of the circulating drift $\vv$ and the entropy production. This is the link the $P_4$ intervention severs: symmetrizing $M$ sets $M_A=0$ and removes this term at a stroke. The first term, the commutator of the \emph{state-dependent} covariance $C(\z)$ with the symmetric part $M_S$, survives symmetrization and is generically nonzero. It is the residual, state-dependent non-integrability noted in Sec.~\ref{sec:results}, and it is why the intervention collapses the entropy production to a small floor rather than exactly to zero, as seen in Figs.~\ref{fig:two} and \ref{fig:three}.

\subsection{The odd-elasticity dictionary}
\label{sec:odd}

The analogy to odd elasticity invoked in Sec.~\ref{sec:local} can now be made
exact. Linearizing the drift about a point, $\f(\z)\approx-K(\z-\z_0)$ with
stiffness $K=-J$, splits as $K=K_S+K_A$ into a reciprocal part $K_S=-J_S$,
derivable from a potential, and a non-reciprocal odd part $K_A=-J_A$. The work
extracted by traversing a closed cycle $C$ in latent space is
\begin{equation}
  W_C=\oint_C \f\cdot\dd\z=\oint_C (J\z)\cdot\dd\z
     =2\,(J_A)_{\perp}\,\mathrm{Area}(C),
\end{equation}
where the symmetric part integrates to zero and, by Stokes, only the component
$(J_A)_\perp$ of $J_A$ in the plane of $C$ survives, times the enclosed area. This
is exactly the odd-elastic work per cycle of a non-reciprocal linear
medium~\cite{scheibner2020}. By Sec.~\ref{sec:irreversibility} the same cycle
integral is the entropy produced, $\Sigma_C=\tfrac1\Dif\oint_C\vv\cdot\dd\z$, and
by the previous subsection $J_A$ descends from the attention asymmetry $M_A$. The
dictionary is therefore
\begin{equation}
\begin{aligned}
  &\underbrace{K_A}_{\text{odd modulus}}
  \;\longleftrightarrow\;
  \underbrace{J_A}_{\text{antisym.\ Jacobian}}
  \;\longleftrightarrow\; \\
  &\underbrace{\vv}_{\text{circulating drift}}
  \;\longleftrightarrow\;
  \underbrace{M_A}_{\text{query--key asymmetry}},
\end{aligned}
\end{equation}
with the odd-elastic work per cycle equal, up to $1/\Dif$, to the entropy produced
per cycle. A reciprocal (symmetric) attention is an ordinary elastic world model
that stores no work around cycles; the odd, non-reciprocal part is supplied by
$W_Q\neq W_K$.

\section{Irreversibility and the query--key asymmetry}
\label{sec:irreversibility}

This section turns the framing of the preceding sections into operational
definitions, so that every quantity is something extracted from a trained world
model rather than assumed. This is what makes the central claim falsifiable, and
it is the part on which the contribution stands or falls.

\subsection{Reversible and irreversible drift from rollouts}

Given an ensemble of latent rollouts, estimate the first two Kramers--Moyal
coefficients locally in $\z$, a standard inference problem for stochastic
trajectories~\cite{frishman2020},
\begin{align}
  \hat\f(\z) &= \frac{1}{\Delta t}\,
    \mathbb{E}\!\left[\z_{t+\Delta t}-\z_t \,\middle|\, \z_t=\z\right],\\
  2\hat\Dif(\z) &= \frac{1}{\Delta t}\,
    \mathrm{Cov}\!\left[\z_{t+\Delta t}-\z_t \,\middle|\, \z_t=\z\right].
\end{align}
The locality (Hooke) assumption of Sec.~\ref{sec:local} is itself testable here:
conditioning additionally on the history $\z_{<t}$ should change $\hat\f$ by only
$O(\epsilon)$, and the size of that change is a direct estimate of $\epsilon$.

Estimate the stationary density $\rho_{\mathrm{ss}}(\z)$ from the occupation of
long rollouts. The steady-state probability current and the irreversible
(circulating) drift are, for constant isotropic $\Dif$,
\begin{equation}
\begin{split}
  \bm{J}(\z) &= \hat\f\,\rho_{\mathrm{ss}} - \hat\Dif\,\nabla\rho_{\mathrm{ss}},\\
  \vv(\z) &= \frac{\bm{J}(\z)}{\rho_{\mathrm{ss}}(\z)}
          = \hat\f(\z) - \hat\Dif\,\nabla\ln\rho_{\mathrm{ss}}(\z),
\end{split}
\label{eq:vmeasure}
\end{equation}
while the reversible part is the gradient
$\hat\f-\vv=\hat\Dif\,\nabla\ln\rho_{\mathrm{ss}}=-\nabla U$ with
$U=-\hat\Dif\ln\rho_{\mathrm{ss}}$. Detailed balance holds if and only if
$\vv\equiv\bm{0}$. Note that $\vv$ and hence the irreversibility are obtained
directly from the current; they do not require having first established the local
Onsager--Machlup form, only the drift and the stationary density.

\subsection{Entropy production}

The path-wise entropy production was defined in Eq.~\eqref{eq:entropy} as
$\Sigma[\pth]=\ln P[\pth]/P[\rpath]$, the central object of stochastic
thermodynamics and its fluctuation
theorems~\cite{seifert2012,jarzynski1997,crooks1999}, which needs only the path
measure and its
reversal. Its steady-state rate is the measurable functional of $\vv$ and
$\rho_{\mathrm{ss}}$,
\begin{equation}
  \dot\Sigma = \int \dd\z\,\frac{|\bm{J}(\z)|^2}{\hat\Dif\,\rho_{\mathrm{ss}}(\z)}
             = \frac{1}{\hat\Dif}\,\big\langle |\vv|^2\big\rangle_{\mathrm{ss}}
             \ge 0 ,
  \label{eq:sigmadot}
\end{equation}
which vanishes exactly when the predicted dynamics is reversible. We propose
$\dot\Sigma$ as the scalar signature of how much genuine temporal structure a
world model predicts.

\subsection{Local circulation and non-normality}

The local generator of circulation is the antisymmetric part
$J_A=\tfrac12(J-J^\top)$ of the drift Jacobian $J$, whose strength we measure by
the frame-invariant non-normality
\begin{equation}
  \mathcal{N}(\z) = \big\|[J,J^\top]\big\|_F = 2\,\big\|[J_A,J_S]\big\|_F ,
\end{equation}
with $J_S=\tfrac12(J+J^\top)$. A drift Jacobian that is symmetric everywhere
($J_A\equiv0$) forces $\vv=\bm{0}$ and hence $\dot\Sigma=0$: such a world model
can only relax to attractors and cannot sustain directed or cyclic prediction.

\subsection{The attention-asymmetry index}

For a transformer world model the attention logit is the bilinear form
$q^\top k = \z_t^\top M\,\z_s$ with $M=W_Q^\top W_K$. The asymmetry under exchange
of the two positions is carried by
\begin{equation}
  M_A=\tfrac12\!\left(W_Q^\top W_K - W_K^\top W_Q\right),
  \qquad
  \mathcal{Q}=\frac{\|M_A\|_F}{\|M\|_F}\in[0,1] ,
\end{equation}
the dimensionless query--key asymmetry. Symmetric attention ($W_Q=W_K$) gives
$\mathcal{Q}=0$; standard attention has $\mathcal{Q}>0$. The bridge proposed here
is that $\mathcal{Q}$ feeds $J_A$ through Eq.~\eqref{eq:vmeasure} and hence drives
$\dot\Sigma$.

\subsection{Falsifiable predictions}

The definitions above make the thesis testable rather than asserted.
\begin{itemize}
  \item[\textbf{P1}] \emph{Locality.} Conditioning the drift on history beyond
        $\z_t$ changes $\hat\f$ by $O(\epsilon)$; this validates the local regime
        and measures $\epsilon$.
  \item[\textbf{P2}] \emph{Broken detailed balance.} $\dot\Sigma>0$ on tasks that
        require predicting sustained dynamics (motion, rhythm, sequence), and
        $\dot\Sigma\to0$ on relaxational tasks (denoising to a fixed point).
  \item[\textbf{P3}] \emph{Correlation.} Across heads, layers, or models,
        $\dot\Sigma$ increases monotonically with the attention asymmetry
        $\mathcal{Q}$.
  \item[\textbf{P4}] \emph{Intervention.} Symmetrizing the attention maps
        ($\mathcal{Q}\!\to\!0$) suppresses $\dot\Sigma$ and selectively degrades
        prediction of sustained dynamics while leaving relaxational prediction
        intact.
  \item[\textbf{P5}] \emph{Instanton versus rollout.} The most-probable path of
        Sec.~\ref{sec:prediction} departs from the deterministic rollout
        $\dot\z=\hat\f$ where $\vv$ and $\nabla\!\cdot\hat\f$ are large.
\end{itemize}
P4 is the causal core: it converts a correlation between an architectural
quantity and a thermodynamic one into an intervention with a predicted,
falsifiable consequence.

\section{Results and discussion}
\label{sec:results}

\subsection{A controlled demonstration}

Before turning to large pretrained world models, we verify both the measurement
pipeline and the mechanism in a controlled minimal model where the architectural
asymmetry can be swept and intervened upon. We take a faithful two-dimensional
attention-driven latent dynamics in which the query--key product
$M=W_Q^\top W_K$ is parameterized as $M=S_0+\theta A_0$, with $S_0=\tfrac12 I$
symmetric and $A_0$ the planar rotation generator, so that the asymmetry
$\mathcal{Q}(\theta)$ runs from $0$ at $\theta=0$ to near unity. We use both a
linear-attention form, for which the latent drift is $\f(\z)=W\z$ with
$W=-\gamma I+M^\top$ and the steady-state entropy production is known exactly
from the Lyapunov equation, and a softmax form
$\f(\z)=-\gamma\z+\beta\sum_j\mathrm{softmax}_j(\z^\top M\bm{m}_j)\,\bm{m}_j$ over
fixed memory slots $\bm{m}_j$. We fix $\gamma=1$ and $\Dif=0.25$; the linear
drift is stable for all $\theta$ (eigenvalues $-\tfrac12\pm i\theta$). The full
pipeline of Eqs.~\eqref{eq:vmeasure}--\eqref{eq:sigmadot} is then applied blindly
to sampled rollouts, with the drift and stationary density estimated on a grid
and never supplied analytically.

\begin{figure*}[t]
  \centering
  \includegraphics[width=\textwidth]{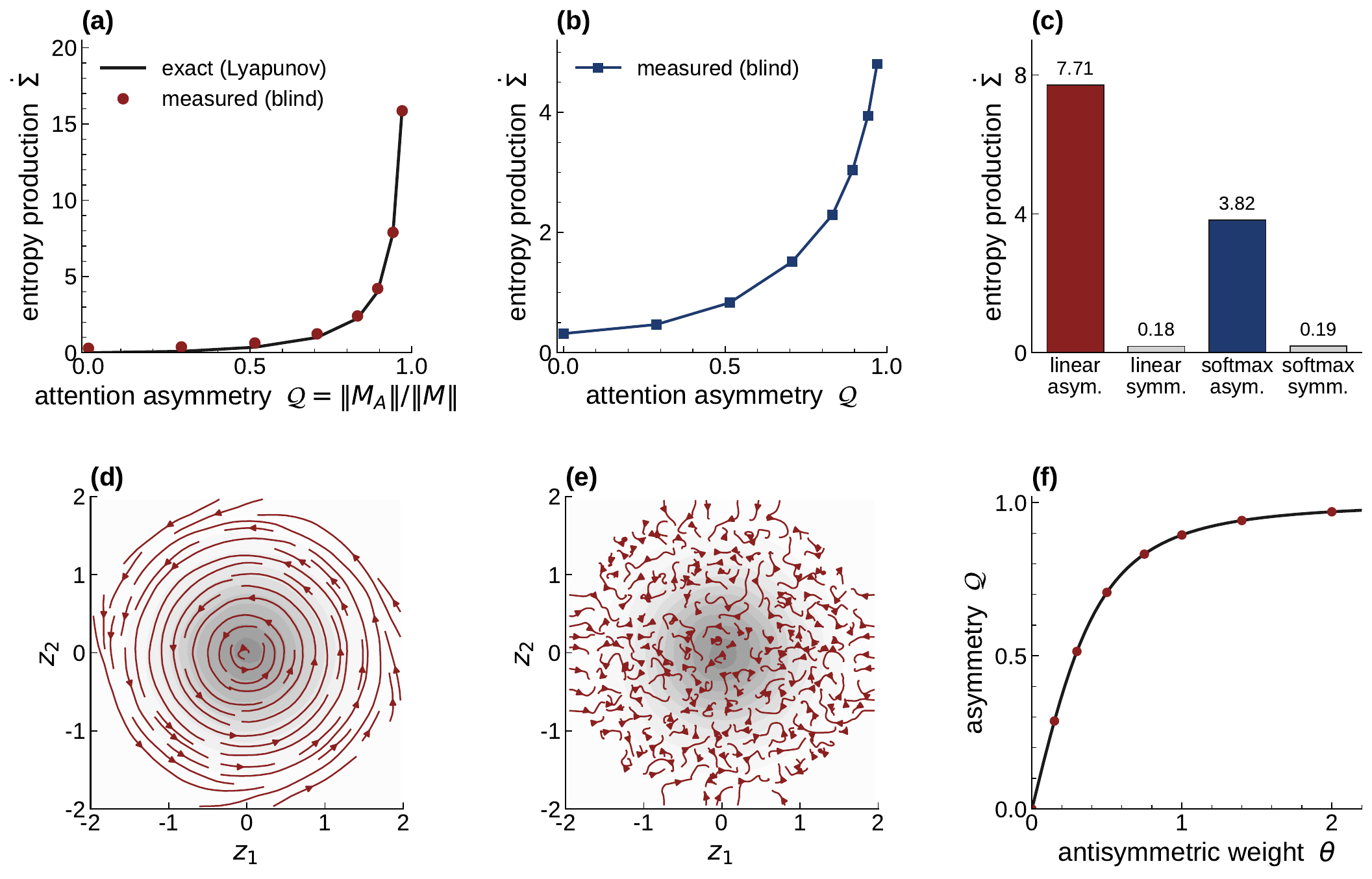}
  \caption{\textbf{Query--key asymmetry sets the entropy production of the
  predicted dynamics.} (a) In a linear-attention latent dynamics, the blindly
  measured entropy production $\dot\Sigma$ rises with the attention asymmetry
  $\mathcal{Q}=\|M_A\|/\|M\|$ and matches the exact Lyapunov value. (b) The same
  trend holds for softmax attention; note the smaller vertical scale, as the
  bounded softmax weights saturate the attainable circulation. (c) Intervention:
  forcing $W_Q=W_K$ ($\mathcal{Q}\!\to\!0$) collapses $\dot\Sigma$ in both models.
  (d,e) The measured probability current (dark red streamlines) over the
  stationary density $\rho_{\mathrm{ss}}$ (gray shading): a coherent global
  circulation for the asymmetric model (d) and no net rotation after
  symmetrization (e), where the residual short vectors are numerical fluctuation
  at the noise floor $\dot\Sigma\!\approx\!0.2$ rather than genuine current.
  (f) The swept asymmetry $\mathcal{Q}(\theta)$; the linear model is stable for
  all $\theta$.}
  \label{fig:one}
\end{figure*}

\subsection{Results}

The four predictions of Sec.~\ref{sec:irreversibility} that the toy model can
address are borne out (Fig.~\ref{fig:one}).

\emph{Pipeline validity and the $\mathcal{Q}$--$\dot\Sigma$ relation (P3).} For
the linear model, the blindly measured $\dot\Sigma$ reproduces the exact Lyapunov
value across the whole sweep (panel a): at $\theta=1$ the exact and measured
rates are $4.00$ and $4.21$, and at $\theta=2$ they are $16.0$ and $15.9$. The
entropy production rises steeply and monotonically with the attention asymmetry,
spanning roughly two orders of magnitude as $\mathcal{Q}$ goes from $0$ to $0.97$.
The agreement establishes that the grid-based Kramers--Moyal and current
estimators recover the true entropy production rather than an artefact of the
measurement.

\emph{Persistence under nonlinearity.} In the softmax model (panel b) the same
monotone rise is observed, from $\dot\Sigma\!\approx\!0.3$ at $\mathcal{Q}=0$ to
$\dot\Sigma\!\approx\!4.8$ at $\mathcal{Q}=0.97$, now saturating at large
$\mathcal{Q}$ because the bounded softmax limits the attainable circulation. The
effect is therefore not an artefact of the linear construction.

\emph{Intervention (P4).} Symmetrizing the attention map, that is forcing
$W_Q=W_K$ so that $\mathcal{Q}\!\to\!0$, collapses the entropy production by more
than an order of magnitude in both models (panel c): from $7.71$ to $0.18$ in the
linear case and from $3.82$ to $0.19$ in the softmax case. Correspondingly, the
measured probability current changes from a coherent global circulation
(panel d) to a field with no net rotation (panel e). Because the intervention
acts on the architecture and the consequence is read out thermodynamically, this
is a causal test of the bridge, not a correlation.

\emph{Noise floor.} The residual $\dot\Sigma\!\approx\!0.2$ that remains at
$\mathcal{Q}=0$ is a positive finite-binning bias of the estimator, the standard
small-sample bias of entropy-production estimators, and it sets the floor of the
measurement. The signal at moderate and large $\mathcal{Q}$ stands well above it.

\subsection{Spontaneous acquisition in a trained model}
\label{sec:fig2}

The toy of Fig.~\ref{fig:one} set the attention asymmetry by hand. To test whether
the same asymmetry is \emph{acquired} by learning, and in proportion to the
irreversibility of the data, we train the attention world model of
Sec.~\ref{sec:attention} on a controllable two-dimensional process
$\dd\z=(-a\z+\omega R\z)\,\dd t+\sqrt{2\Dif}\,\dd\bm W$, with $R$ the rotation
generator, whose true entropy production is $\dot\Sigma_{\mathrm{true}}=2\omega^2/a$.
The model queries the state against learnable memory slots with values tied to
keys, so its only channel for an antisymmetric drift is the score matrix
$M=W_Q^\top W_K$; we initialize it symmetric ($\mathcal{Q}=0$) and train by
next-step prediction.

\begin{figure*}[t]
  \centering
  \includegraphics[width=\textwidth]{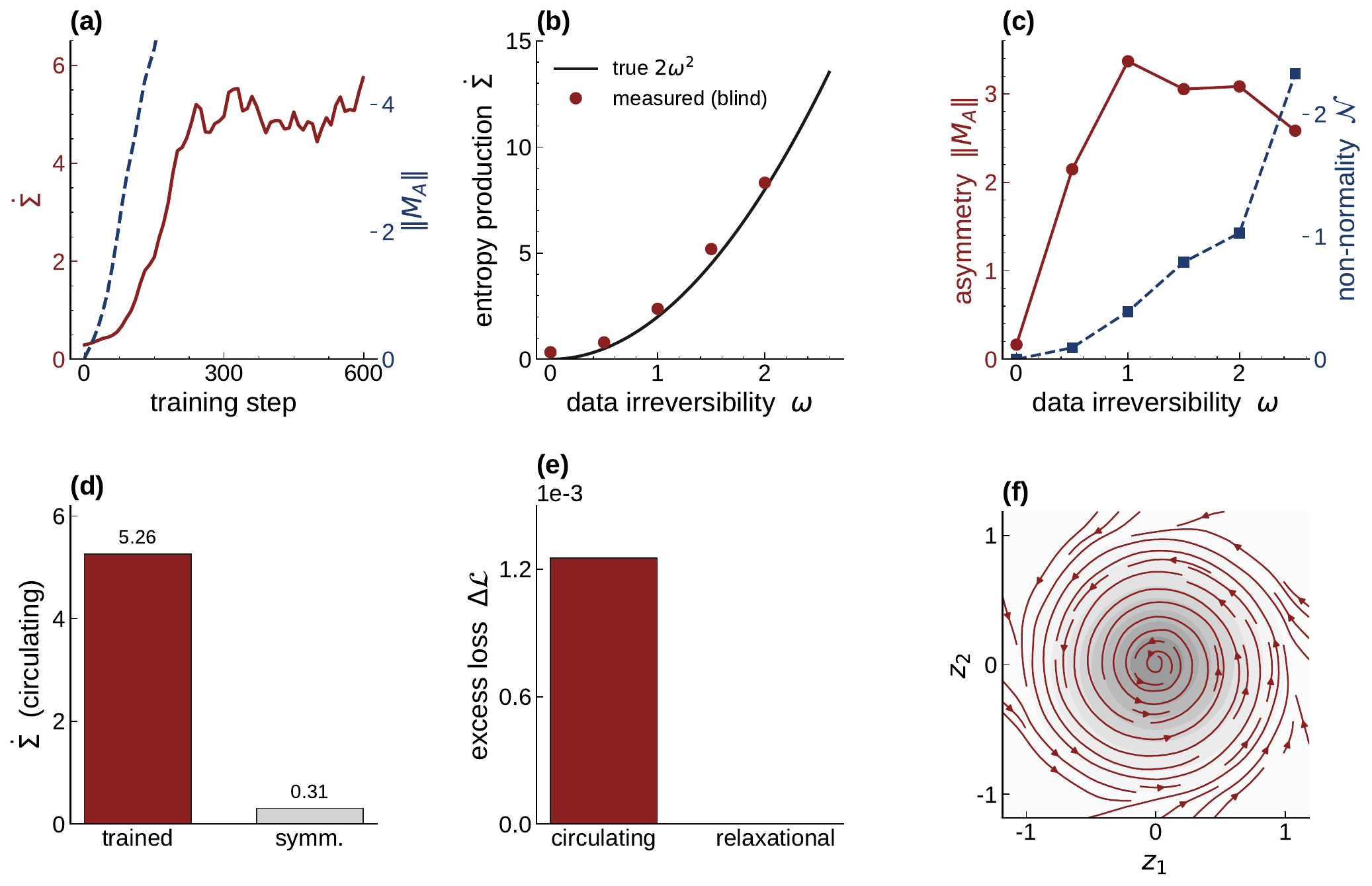}
  \caption{\textbf{Query--key asymmetry is acquired by training in proportion to
  the data's irreversibility.} (a) From a symmetric initialization, training on
  circulating data drives $\|M_A\|$ (dashed) and the measured $\dot\Sigma$ (solid)
  up together. (b) The trained model's blind $\dot\Sigma$ reproduces the true
  $2\omega^2/a$ across the data irreversibility $\omega$. (c) The acquired
  asymmetry $\|M_A\|$ (left axis) and the dynamical non-normality
  $\mathcal{N}=\|[J,J^\top]\|_F$ it induces (right axis) both vanish for gradient
  data ($\omega=0$) and grow with circulating data; the ratio $\mathcal{Q}$
  saturates and $\|M_A\|$ turns over at
  large $\omega$, where the model uses additional capacity. (d) Symmetrizing the
  learned $M$ collapses $\dot\Sigma$ (and, in the text, $\mathcal{N}$). (e) Excess
  one-step prediction loss after
  symmetrization: circulating prediction degrades, relaxational prediction is
  spared. (f) The trained circulating model's measured probability current (dark
  red) over its stationary density (gray shading).}
  \label{fig:two}
\end{figure*}

Figure~\ref{fig:two} reports the outcome. Starting from a symmetric initialization,
training on circulating data drives the query--key asymmetry $\|M_A\|$ and the
measured entropy production up together within the first hundred steps (panel a):
the asymmetry is acquired, not imposed. Across the sweep the trained model's blind
entropy production reproduces the true $2\omega^2/a$ (panel b), and the acquired
asymmetry, together with the dynamical non-normality $\mathcal{N}$ it induces, is
smallest for gradient ($\omega=0$) data, where the drift Jacobian is exactly
normal, and grows with the circulating part of the data (panel c). The causal test
is decisive: symmetrizing the learned $M$ collapses the
entropy production from $5.0$ to $0.3$ (panel d), and with it the non-normality
$\mathcal{N}$ from $0.8$ to $0.06$, and selectively raises the
prediction loss on circulating data while leaving relaxational data untouched
(panel e), and the trained model's measured current is a coherent circulation
(panel f). A small attention model thus learns to be irreversible exactly when its
data is, and carries that irreversibility in its query--key asymmetry.

\subsection{Irreversibility is a resource for prediction}
\label{sec:fig3}

The measurements so far show that a trained model is irreversible and that its
irreversibility is carried by the query--key asymmetry. The sharper question is
whether that irreversibility is \emph{useful}: does removing it cost prediction? We
compare multi-step deterministic rollouts of the trained model and of its
symmetrized counterpart against the true conditional mean
$\mu(\z,h)=e^{W_\omega h}\z$, on circulating and relaxational data.

\begin{figure*}[t]
  \centering
  \includegraphics[width=\textwidth]{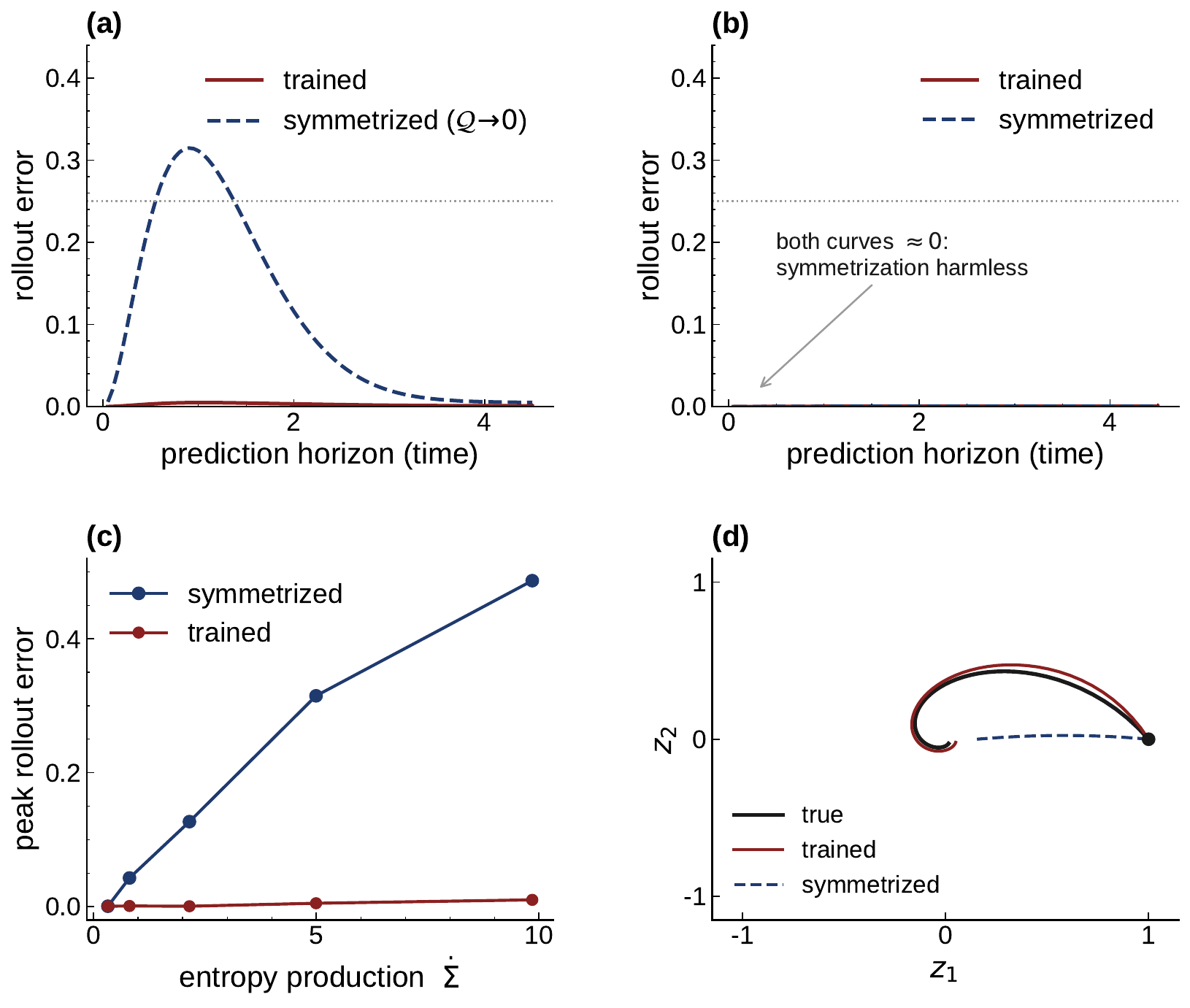}
  \caption{\textbf{Irreversibility is a resource for long-horizon prediction.}
  Multi-step rollout error against the true conditional mean. (a) On circulating
  data the trained model (solid) tracks the truth while the symmetrized model
  (dashed, $\mathcal{Q}\!\to\!0$) fails at long horizon. (b) On relaxational data
  the two coincide, so symmetrization is harmless. (c) The peak prediction error
  caused by symmetrization grows monotonically with the entropy production
  $\dot\Sigma$, while the trained model stays near zero. (d) An example circulating
  rollout: the trained model follows the true spiral, the symmetrized model decays
  radially and loses the rotation.}
  \label{fig:three}
\end{figure*}

Figure~\ref{fig:three} answers it. On circulating data the trained model tracks the
true rollout over the full horizon, while the symmetrized model's error grows
sharply: unable to rotate, it predicts only radial relaxation and loses the phase
(panels a, d). On relaxational data the two are indistinguishable (panel b), so
symmetrization costs nothing when there is no circulation to predict. Across the
sweep, the prediction lost on symmetrization grows monotonically with the entropy
production (panel c). The query--key asymmetry is therefore not a passive
by-product but a resource: the entropy it produces is what lets the model predict
irreversible structure over a long horizon, and removing it degrades exactly that
prediction and nothing else. A world model that must anticipate genuine temporal
structure cannot be at equilibrium.

\subsection{Discussion}

The demonstration confirms, in a controlled setting, the paper's central claim:
the irreversibility of a world model's predicted dynamics is set by the asymmetry
of its attention maps. The reading is physical. A symmetric attention
($W_Q=W_K$) yields a gradient drift, the equilibrium, modern-Hopfield limit, in
which the latent state merely relaxes to an attractor and the predicted dynamics
is time-reversible. Only an asymmetric attention produces the circulating drift
$\vv$ that sustains directed or cyclic prediction, and that circulation is
exactly the entropy production measured here. A predictive model that must
anticipate ongoing temporal structure, motion, rhythm, or sequence, therefore
cannot operate at equilibrium: it must break detailed balance, and the resource
that lets it do so is supplied architecturally by the query--key asymmetry. This
gives a mechanistic counterpart to, and is consistent with, the empirical finding
that neural and active biological dynamics break detailed balance, more strongly so
under demanding cognitive tasks~\cite{lynn2021,battle2016,gnesotto2018}, and with
the view of the brain as a prediction machine that minimizes a free-energy
functional~\cite{friston2010}; the difference is that here the irreversibility
is traced to an identifiable architectural source rather than only measured in the
data. Figure~\ref{fig:three} turns this from an argument into a measurement:
removing the asymmetry destroys precisely the long-horizon, circulating prediction
it was claimed to support and leaves relaxational prediction intact, so the entropy
production is a resource the predictor uses rather than waste heat it merely
emits~\cite{parrondo2015}.

Several limitations bound the claim. The model is a two-dimensional toy rather
than a trained world model, chosen so that $\mathcal{Q}$ can be swept and
intervened upon and so that an exact cross-check exists. What keeps the
demonstration from being a two-dimensional coincidence is that the mechanism behind
it is established analytically and independently of dimension:
Eq.~\eqref{eq:JAfromMA} expresses the antisymmetric drift Jacobian through the
query--key asymmetry $M_A$ for any value-tied attention layer in any latent
dimension, so the experiment confirms a general algebraic relation rather than a
property of the particular system. What remains genuinely scale- and
architecture-dependent, and is therefore the proper subject of a large-model study,
is whether the query--key channel keeps dominating the value path once values are
untied, and whether a deep stack preserves the local-action picture. We are careful
to claim that Eq.~\eqref{eq:JAfromMA} \emph{identifies and controls} a channel of
irreversibility, not that this channel \emph{dominates} it in a full architecture:
residual connections, the feedforward block, layer normalization, value mixing, and
depth composition all contribute to the learned drift, and establishing which
channel dominates the entropy production of a large pretrained model is the open
empirical question, separate from the analytic identification made here. The floor of $\dot\Sigma\!\approx\!0.2$ limits the dynamic range at small $\mathcal{Q}$. 
This floor receives contributions both from finite-binning bias and from the state-dependent residual non-integrability of Sec.~\ref{sec:results}; separating the two would require comparing floors across binning resolutions, which we do not pursue here.
The linear panel makes the $\mathcal{Q}\!\to\!\dot\Sigma$ link nearly analytic, which is why the softmax panel is included as a guard against a purely linear artefact.
Finally, the grid-based current estimator used here does not scale to the
high-dimensional latents of real models. Three routes avoid the grid. The drift
$\f$, and hence $\vv$ through Eq.~\eqref{eq:vmeasure}, can be estimated by neural
score matching of the one-step transition~\cite{hyvarinen2005,song2021,ho2020},
since $\nabla\ln\rho_{\mathrm{ss}}$ is
a score of the kind such networks already learn. The rate $\dot\Sigma$ can be
estimated directly with a variational lower bound on the divergence between the
forward and reversed path distributions, which needs only samples and no
explicit density, in the spirit of recent machine-learning estimators of entropy
production~\cite{barato2015,otsubo2020,kim2020}. Or $\Sigma[\pth]$ can be accumulated
pathwise from the model's
own trajectory log-likelihoods. All three reduce to the quantities of
Sec.~\ref{sec:irreversibility} and are what we would use on a trained model.

A further subtlety concerns the intervention. Symmetrizing $M=W_Q^\top W_K$
removes the asymmetry at the level of the weights, but the softmax makes the drift
state-dependent, and a state-dependent map can be non-integrable even under a
symmetric $M$: the antisymmetric part of the drift Jacobian contains a term
$\propto[\langle a\,\bm{m}\bm{m}^\top\rangle(\z),\,M]$ that need not vanish when
$M=M^\top$. Our claim is therefore about the weight-level channel, the resource a
designer controls through $W_Q,W_K$, and the empirical collapse of $\dot\Sigma$ to
the floor under symmetrization (Figs.~\ref{fig:two},~\ref{fig:three}) shows that
this channel dominates here. A complete account would also measure the residual
state-dependent non-integrability under symmetric weights, which we leave to the
high-dimensional study above.

Figure~\ref{fig:two} takes the first of these steps on a small trained model. The
remaining rung is a large pretrained latent world model, an RSSM or Dreamer-type
model or a transformer trained on rich data, where the latent is high-dimensional
and the grid estimator gives way to the score-matching and variational estimators
above. 
There the value path is no longer tied, so the analysis must also ask how the irreversibility is shared between the query--key and value paths, and the selective degradation should be read over long-horizon and planning rollouts, where one-step effects compound, rather than the single step measured here. 
The learned drift is then also shaped by the non-attention channels: the residual connections, feedforward blocks, and normalization. Their contributions can be isolated by comparing the measured drift against one in which those blocks are bypassed or ablated, so that the query--key channel is read against, rather than confounded with, the rest of the architecture.
The operational definitions of Sec.~\ref{sec:irreversibility} carry over unchanged; only the estimators and the scale grow.

Concretely, the protocol on a trained model is the following. Extract
$\mathcal{Q}=\|M_A\|/\|M\|$ for each head and layer from its $W_Q,W_K$; measure
$\dot\Sigma$ of its latent rollouts with the estimators above; and test the
correlation (P3) across heads, layers, and training checkpoints. The causal test
(P4) then symmetrizes $W_Q^\top W_K\to\tfrac12(W_Q^\top W_K+W_K^\top W_Q)$ and
predicts three coupled consequences: $\dot\Sigma$ falls; long-horizon and
periodic-motion prediction, for example physics-engine rollouts, degrades; and
planning quality drops; while purely relaxational tasks, such as denoising or
completion to a static structure, are spared. This selective degradation,
irreversible prediction lost but relaxation preserved, is the signature that
would turn the central claim from a hypothesis into a measured result.

\section{Conclusion}

In this work, we proposed a path-space formulation of prediction in AI world
models. Rather than viewing prediction as the estimation of a future state
distribution, we argued that a world model implicitly represents a probability
distribution over future trajectories. Within this framework, prediction,
planning, and uncertainty emerge as different operations on the same path measure:
the most probable trajectory corresponds to prediction, constrained trajectory
selection corresponds to planning, and fluctuations around dominant trajectories
correspond to uncertainty.
Assuming an effective local path description, the trajectory distribution takes the
form of an Onsager--Machlup action. This provides a common physical language
linking latent dynamics, stochastic thermodynamics, and attention-based sequence
modeling. The resulting framework allows reversible and irreversible components of
the learned dynamics to be separated, making entropy production a measurable
property of the predicted world rather than an abstract thermodynamic quantity.
Building on this formulation, we introduced operational measures connecting
architectural asymmetry in attention mechanisms to the irreversibility of learned
dynamics. In controlled attention-model experiments, asymmetry in the attention
structure was associated with circulating probability flow and positive entropy
production, while symmetrization interventions selectively degraded long-horizon predictive performance. These observations support the hypothesis that irreversibility is not only a by-product of learning but may constitute a computational resource for maintaining predictive representations of temporally
evolving environments. 

Future prediction is naturally formulated as inference on path space, within which irreversibility emerges as a measurable property of the predictive dynamics and, potentially, as a computational resource for representing sustained temporal structure. A world model, in this view, does not predict the next state; it weights whole futures, and its irreversibility is what makes the long-horizon structure of those futures predictable.
The present work should be viewed as a first step toward a statistical physics of
world models. The central object is not a state distribution but a path
distribution, and many questions remain open. In particular, it remains to be
established to what extent large-scale world models admit an effective local path
description, how nonlocal memory effects modify the trajectory measure, and whether
the thermodynamic structures identified here persist across architectures and
scales. We hope that the path-space perspective developed in this work provides a
foundation for addressing these questions and for building a unified physical
theory of prediction, planning, and uncertainty in intelligent systems.

\end{document}